\documentclass[11pt]{article}

\usepackage[final]{acl}

\usepackage{times}
\usepackage{latexsym}
\usepackage[T1]{fontenc}
\usepackage[utf8]{inputenc}
\usepackage{microtype}
\usepackage{inconsolata}
\usepackage{booktabs}
\usepackage{amsmath}

\title{The Missing ``I Don't Know'': Why Three Reasoning-Reliability Findings Converge on Calibrated Abstention}

\author{Srijith Ravikumar\thanks{This work is independent of the author's role at Amazon and was
conducted without the use of Amazon proprietary data, systems, or resources. The views expressed
are the author's own.} \\
  Amazon.com LLC \\
  \texttt{srijith@amazon.com}}

\begin{document}
\maketitle

\begin{abstract}
Three recent results describe what look like unrelated LLM reliability problems. \citet{yin2025reasoning} show reasoning RL collapses tool-reliability representations. \citet{suleymanov2026beyond} show that under safety-constrained generation, large models rewrite flagged spans while small models truncate. \citet{bastounis2024consistent} prove any consistent-reasoning system without an implicit ``I don't know'' function must hallucinate infinitely often on broad problem classes. We argue these findings converge on a single intervention: calibrated abstention is what each independently identifies as the missing capability, even though the unavailability they document --- a capability gap, a policy gap, and a recursion-theoretic gap --- has a different source in each case. Honesty post-training has narrowed the gap in deployed models, but principled closure of the class Bastounis identifies requires an implicit ``I don't know'' function, and the leaderboard signal that would select for its practical analogue is missing: dominant benchmarks assign zero reward to decline, so the leaderboard gradient that would select for the function does not exist where model releases are ranked. We propose four changes to evaluation: triple-scoring, abstention-rate reporting, capability-stratified evaluation, and mandatory calibration metrics. Benchmark reform is necessary, not sufficient, for closing the gap the theorem identifies.
\end{abstract}

\section{Introduction}

Three results from the past two years describe what looks like three different problems. \citet{yin2025reasoning} show that reinforcement learning for reasoning collapses the internal representations that govern tool reliability, producing agents that invoke tools they know are unavailable. \citet{suleymanov2026beyond} show that when an agentic loop constrains outputs away from sensitive completions, large models rewrite affected spans into nuanced alternatives while small models truncate text to the point of non-utility. \citet{bastounis2024consistent} prove formally that any system that reasons consistently across paraphrased equivalents must hallucinate infinitely often on a broad class of problems, including basic arithmetic, unless it computes an implicit ``I don't know'' function.

We argue these findings converge on a single intervention. The unavailability each documents has a different source --- a capability gap (Yin), a policy gap (Suleymanov), and a recursion-theoretic gap that is agnostic to source (Bastounis) --- but in each case the failure mode is what happens when a system without calibrated abstention is forced to commit on inputs where commitment is unwarranted. Yin and Suleymanov are not instances of Bastounis's theorem; they are independent empirical settings whose failure modes are most parsimoniously addressed by the same construction the theorem identifies. We treat the three results as convergent evidence for calibrated abstention rather than as instances of a single structural phenomenon, and argue in Section~\ref{sec:unification} that the convergence has policy consequences the field has not yet acted on. The reading is defeasible from the start: it fails if a mechanism-level intervention implementing no abstention or selective-prediction construction closed all three failure modes at once. Section~\ref{sec:unification} states that condition precisely.

This matters because most work on the reliability problem has treated it as a tuning problem. Better RL objectives, better prompting, better alignment. Honesty post-training has narrowed the gap in deployed models. But the theorem rules out closure in principle for any consistent-reasoning system that lacks the implicit ``I don't know'' function, and current training pipelines do not compute it at the structural level the theorem requires. A parallel line of abstention-specific work \citep{kamath-etal-2020-selective,wen-etal-2025-know,kirichenko2025abstentionbench} has developed mechanisms but has not produced reform of the dominant leaderboard benchmarks, which is where model releases are selected. What the theorem requires for closure is the implicit ``I don't know'' function, and the selection signal for its practical analogue is currently backwards.

Two claims follow. First, the three findings converge on calibrated abstention as the warranted intervention, and treating them separately has obscured that convergence. Second, benchmark reform is necessary --- though not sufficient --- for closing the reasoning-reliability gap that the theorem identifies. Honesty post-training partially substitutes for the missing signal; principled closure requires that benchmarks stop assigning zero reward to decline. We propose triple-scoring across correct, wrong, and abstained outcomes, abstention-rate reporting, capability-stratified evaluation, and calibration metrics (ECE, Brier) alongside accuracy.

\section{The Three Findings}

\subsection{Reasoning RL collapses tool-reliability representations}

\citet{yin2025reasoning} build SimpleToolHalluBench, which tests two failure modes where abstention is the correct behavior: queries where no tool is available and queries where only a distractor tool is present.

They find that reinforcement learning targeting reasoning raises tool hallucination, and that the rise is systematically associated with reward-signal gains. The effect holds even when the RL training set contains no tools: GSM8K math fine-tuning increases hallucination on tool benchmarks the model never saw during training. The effect appears both under GRPO-style outcome-centric RL and when reasoning is only elicited at inference time by prompt switching. The mechanism is a representation collapse: tool-reliability pathways destabilize in early and middle layers while general reasoning pathways stay stable, and activation probes localize the resulting errors as divergences concentrated in late-layer residual streams. Mitigations fail cleanly. Prompt instructions give marginal gains. DPO reduces hallucination only by sacrificing utility. The authors call this a reliability-capability trade-off with no known resolution.

\subsection{Safety-constrained generation produces scale-dependent failure modes}

\citet{suleymanov2026beyond} study what happens when an agentic self-correction loop (SemSIEdit) flags and rewrites spans containing sensitive inferences. Across 13 models they document the rewrite-versus-truncate divergence. GPT-5 responds with constructive expansion, increasing mean answer length by 110 characters (Suleymanov et al.'s headline figure across the GPT-5 condition) as it rewrites flagged spans into nuanced alternatives while keeping utility at 8.58/10. Smaller and mid-sized models respond with destructive truncation, reducing length by 50--73\% across the smaller-model conditions reported. Both strategies reduce leakage; only the first preserves utility. The authors argue the Editor role requires a minimum reasoning threshold, below which agentic refinement collapses into refusal. The bimodality is presented at the headline level; we cite the headline numbers and do not infer significance beyond the source paper's own qualifications.

A second finding inverts the direction. Toggling chain-of-thought on Qwen3-8B with no defense raises leakage from 74\% to 84\% (single-run point estimates as reported): reasoning lets the model connect benign facts into sensitive inferences. Running the same model with SemSIEdit active is consistent with sign inversion in the reported numbers, dropping leakage to 42\% versus 47\% without reasoning (single-run point estimates). The capability that amplifies the failure mode unconstrained is the same capability that powers the rewrite under constraint.

\subsection{Consistent reasoning without abstention is provably unreliable}

\citet{bastounis2024consistent} prove the Consistent Reasoning Paradox. Consistent reasoning, in their formal sense, is a property of a system implementable as a Turing machine: correctness on a task propagates across a computably enumerable family of paraphrased equivalents of it (``Tell me the time!'' and ``What is the time?''). The theorem establishes that any such system that always answers will hallucinate infinitely often on certain problem classes, and that basic arithmetic contains such a class. The escape they identify is an implicit ``I don't know'' function: a partial computable refusal that fires on the subset of inputs where consistent reasoning would force confabulation. They state the requirement strongly. Building an AI that is trustworthy, meaning one that never answers incorrectly, and that also reasons consistently, ``can only be done by implicitly computing \ldots{} the `I don't know' function,'' and computing it in turn requires a giving-up parameter rather than a boundary fixed in advance. What they do not claim is that practical calibrated abstention --- a softmax-thresholded refusal head, a conformal selection layer \citep{quach2024conformal}, or a Kadavath-style $P(\mathrm{True})$ self-evaluator \citep{kadavath2022language} --- instantiates that recursion-theoretic object. We adopt the I-don't-know function as a target the missing benchmark gradient should select for, and invoke the theorem as the source of a convergent failure mode rather than applying it to LLM inference directly, since we do not claim current models achieve the consistency property it assumes.

Each of the three papers stands on its own. Yin's work is a mechanistic empirical study. Suleymanov's is behavioral. Bastounis's is a theorem in computability theory. Taken separately they explain different phenomena. We argue in Section~\ref{sec:unification} that they converge on the same intervention.

\section{Convergent Evidence}
\label{sec:unification}

The three findings converge on the same intervention. Each documents what happens when a system that lacks calibrated abstention is forced to commit on inputs where commitment is unwarranted, but the source of the unwarranted commitment differs across the three. Yin documents a capability gap: the model lacks the tool, the deductive path, or the fact, and outcome-reward RL has selected against the only honest response. Suleymanov documents a policy gap: the surrounding agentic loop has flagged the natural completion as sensitive, leaving the model to produce something else in its place. Bastounis proves the recursion-theoretic case: even an idealized consistent-reasoning system must hallucinate infinitely often on inputs where the implicit ``I don't know'' function would fire. Three different unavailabilities, three different empirical regimes, one shared remedy: the abstention construction the theorem identifies as the escape from its impossibility result. Four distinct constructs travel under the name abstention in this literature, and the reform we propose scores only one of them. Table~\ref{tab:taxonomy} separates them, and we keep them separate throughout.

We use ``convergent evidence'' deliberately, and not ``one phenomenon'' or ``a single structural property.'' The theorem applies to systems achieving recursion-theoretic consistency over computable paraphrase classes; GRPO selects on outcome reward over rollouts of the same prompt and does not, in general, achieve that property. The simpler claim suffices: outcome-reward RL with no abstention reward selects against decline across the entire training distribution, the model is therefore optimized to commit on inputs where commitment is unwarranted, and what closes that gap in principle is the I-don't-know construction the theorem identifies. The convergence is at the level of the missing capability, not at the level of the formal property each system achieves. The reading is defeasible, and we name the condition that would defeat it: the convergence is refuted if a mechanism-level intervention implementing no abstention or selective-prediction construction closed all three failure modes at once: Yin's tool hallucination, Suleymanov's truncation under constraint, and the reasoning-induced abstention degradation \citet{kirichenko2025abstentionbench} measure. Bastounis's result is a theorem and cannot be empirically closed, so the third target is its measurable manifestation rather than the formal result.

\begin{table*}[t]
\centering
\small
\setlength{\tabcolsep}{6pt}
\renewcommand{\arraystretch}{1.2}
\begin{tabular}{@{}p{0.22\textwidth}p{0.34\textwidth}p{0.36\textwidth}@{}}
\toprule
\textbf{Construct} & \textbf{Fires when} & \textbf{Where it is scored} \\
\midrule
Recursion-theoretic ``I don't know'' & Consistent reasoning would force confabulation & Target of the missing gradient; not directly measured (Section 2.3) \\
Epistemic abstention & Calibrated confidence falls below $p^*$ & Triple-scoring plus a proper score on commitments (Item~1) \\
Prescriptive or safety refusal & Policy rules out the natural completion & Safety benchmarks, separately from capability \\
Partial or contextual answer & Commitment unwarranted, partial information available & Open for long-form and interactive settings (Limitations) \\
\bottomrule
\end{tabular}
\caption{Four constructs this paper keeps distinct. Only the second is scored by the proposal of Section~\ref{sec:proposal}: the first is the theorem's formal object and not directly measurable, the third belongs to safety evaluation and would confound capability scoring if folded in, and the fourth is what keeps the reform scoped to discrete answers.}
\label{tab:taxonomy}
\end{table*}

This frame absorbs Yin without invoking the theorem mechanistically. Reasoning RL on outcome rewards selects against partial answers and honest failures relative to coherent commitments. Running the training on GSM8K still amplifies tool hallucination because what RL is teaching is not tool behavior; it is the policy of always producing a numeric output. When the evaluation distribution shifts to queries where the correct response is decline, the policy generalizes anyway. Late-layer residual divergence is the activation-space signature of a model trained to produce numeric outputs producing them where it should not. \citet{kirichenko2025abstentionbench} build an abstention-specific benchmark across 20 datasets and report that reasoning fine-tuning degrades abstention by 24\% on average (their headline figure, averaged over the two reasoning and non-reasoning model pairs they compare), including in domains the reasoning models were explicitly trained on. The empirical pattern is consistent with what the theorem requires: a system without the I-don't-know function commits where decline would be correct. Whether the precise mechanism is recursion-theoretic consistency-forcing or simpler selection-against-decline, the proposed remedy is the same.

This resolves an apparent tension between Kirichenko and Suleymanov. In Kirichenko's setting, expanded test-time compute worsens abstention as the model uses extra tokens to manufacture missing context. In Suleymanov's, expanded compute enables the constructive rewrite. Reasoning capability is dual-use: its impact on reliability depends on whether the surrounding loop rewards honest refusal or mandates a definitive output. The two settings differ in their loops, not in their reasoning.

Suleymanov fits the same convergence from the opposite direction and tells us what varies with capability. SemSIEdit creates regions where the model's natural completion has been flagged as sensitive. The model must produce something in those regions; what it produces depends on whether it can compute a local ``I don't know, here is what I can still say'' construction at each flagged span. GPT-5 can. The +110 character expansion that \citet{suleymanov2026beyond} report (a headline mean across the GPT-5 condition) is the empirical shape of an abstention-with-context intervention rather than a measured calibrated abstention in the technical sense. Gemma-3-4B cannot, and so cuts rather than rewrites. Suleymanov's setting is technically prescriptive refusal (the policy rules out the natural completion), distinct from \citet{kirichenko2025abstentionbench}'s epistemic-abstention frame. What is shared is the requirement itself: in both cases the model must interrupt a probable token sequence and substitute a calibrated alternative. Suleymanov's setting therefore enters the convergence as evidence that reasoning capability is dual-use, not as a benchmark the reform would rescore; the policy gap is where safety benchmarks already operate.

The same reasoning capacity that amplifies leakage when no constraint is active produces safer output when the constraint is active, because the model can now spend compute on the rewrite rather than on connecting facts into sensitive inferences. Reasoning without abstention infrastructure is what Yin documents. Reasoning with the beginnings of an abstention infrastructure is what Suleymanov's GPT-5 result looks like. Bastounis's theorem tells us why the distinction is structural, not incidental.

\section{Why Current Benchmarks Make This Worse}

The four benchmarks that dominate LLM leaderboards (MMLU \citep{hendrycks2021measuring}, HellaSwag \citep{zellers2019hellaswag}, HumanEval \citep{chen2021humaneval}, GSM8K \citep{cobbe2021gsm8k}) all share a property. None provides an abstention option with nonzero reward. Table~\ref{tab:benchmarks} shows the structure. Across all four, expected reward for any attempt weakly exceeds reward for decline. For the two multiple-choice rows the inequality is strict and guessing dominates even with no signal; for the open-generation rows it follows whenever the model has any signal above noise. A model optimized against these leaderboards learns the policy the scoring function rewards.

\begin{table*}[t]
\centering
\small
\setlength{\tabcolsep}{8pt}
\renewcommand{\arraystretch}{1.2}
\begin{tabular}{@{}lllc@{}}
\toprule
\textbf{Benchmark} & \textbf{Output format} & \textbf{Attempt reward} & \textbf{Abstain reward} \\
\midrule
\multicolumn{4}{@{}l}{\emph{Multiple-choice (random guess strictly dominates abstain):}} \\
MMLU & 4-choice MCQ & 0.25 (uniform guess) & 0 \\
HellaSwag & Sentence-completion MCQ & 0.25 (uniform guess) & 0 \\
\midrule
\multicolumn{4}{@{}l}{\emph{Open-generation (any output rewarded over none):}} \\
HumanEval & Unit-test pass@$k$ & $\geq 0$, $>0$ in expectation & 0 \\
GSM8K & Exact-match numeric & $\geq 0$, $>0$ in expectation & 0 \\
\bottomrule
\end{tabular}
\caption{Expected scores under four dominant LLM benchmarks. Reward for any attempt weakly exceeds reward for decline in every row. The argument is sharper for multiple-choice (random guessing strictly dominates abstention regardless of model knowledge) than for open-generation (where the inequality follows by construction whenever the model has any signal above noise).}
\label{tab:benchmarks}
\end{table*}

Yin's GSM8K-to-tool-hallucination result (Section 2.1) is what this incentive structure produces downstream when the benchmark is also used as an RL training signal. GRPO's group-relative advantages are computed from outcome rewards on GSM8K answers, and those rewards are zero for any response that fails to produce a numeric match. No reward signal touches ``decline.'' The model is selected for producing numeric outputs whether or not they are correct, and the policy generalizes beyond the training distribution to evaluation tasks where decline would be the correct response. We distinguish three levels and keep them separate. The \emph{training-time gradient} is the RL signal GSM8K-as-RL-data exerts, and it is what produces Yin's downstream effect. \emph{Evaluation-time selection} is post-hoc selection on model releases, which all four benchmarks in Table~\ref{tab:benchmarks} exert and which Section~\ref{sec:proposal} targets. The \emph{deployment-time product decision} is where the actual-use $R_{\text{incorrect}}$ asymmetry of medical or legal deployment is felt. The reform acts on the second, and on the first only where a benchmark doubles as training data. It does not fix the third --- that is where honesty post-training operates.

Error-rate audits compound the problem. \citet{gema-etal-2025-mmlu} find a 6.49\% mean error rate across MMLU questions, rising to 57\% in the Virology subset (where the small denominator widens the Wilson-interval bound considerably). A calibrated model that refuses to commit on an ambiguous or mislabeled item scores identically to a model that commits to the labeler's mistake. The benchmark cannot distinguish appropriate abstention from failure.

These four also share a format: every one is discrete-answer, whether multiple choice, exact match, or unit-test pass. The discrete-answer scope of the proposal below is therefore not a convenience. It is where the selection signal is actually set. A model with no gradient pointing toward ``I don't know'' is what the three papers of Section 2 document.

\section{A Concrete Proposal}
\label{sec:proposal}

Prior work has developed the mechanisms we invoke and several adjacent benchmark reforms. The selective-prediction literature \citep{geifman2017selective,kamath-etal-2020-selective,el2010foundations} provides the risk-coverage framework that Equation~\ref{eq:reward} below specializes to a three-action benchmark setting. Mechanism-level abstention work has produced semantic-uncertainty estimates \citep{kuhn2023semantic}, refusal-tuning \citep{zhang2024rtuning}, conformal generation \citep{quach2024conformal}, self-knowledge calibration \citep{kadavath2022language}, and training-time ternary-reward RL \citep{wei2025truthrl,zhai2026abstainr1,ralethe2025kerlqa} that optimizes three-way rewards involving abstention directly during model training, surveyed broadly in \citet{wen-etal-2025-know}. Existing benchmark reforms address adjacent dimensions of the same problem: HELM \citep{liang2022holistic} introduced multi-metric leaderboards including calibration; TruthfulQA \citep{lin2022truthfulqa} credits truthful non-commitment without scoring decline as a distinct action; AbstentionBench \citep{kirichenko2025abstentionbench} measures abstention directly; FActScore \citep{min2023factscore} decomposes factuality. What has not been argued is that these reforms compose into a structural requirement, not an optional companion to accuracy. We ground necessity in the convergent-evidence argument of Section~\ref{sec:unification} rather than in a direct application of the theorem, since current LLMs do not strictly inhabit its formal class. The contribution below is that claim and the specific bundle, not the individual reforms. \citet{wei2025truthrl}, \citet{zhai2026abstainr1}, and \citet{ralethe2025kerlqa} use closely related training-time three-way rewards, but all operate on the training objective, and the dominant evaluation pipelines stay broken even where a training pipeline has adopted one. Without leaderboard-side reform those advances stay isolated rather than compounding.

Recent work has begun to operationalize parts of this. \citet{kalai2025hallucinate} call for rescoring the benchmarks that dominate leaderboards and propose auditing behavior across explicit confidence targets; \citet{jackson2025omniscience} run a bounded index that credits abstention at a fixed penalty of $\lambda = 1$ and report only three of thirty-six models scoring above zero; \citet{mohamadi2025hesitation} train on the ternary reward directly; \citet{cheng2025facts} maintain a live factuality leaderboard reporting hedging rate beside accuracy. Four independent efforts inside a year is evidence for the position argued here. None of them stratifies by capability or audits abstention across subgroups, and none treats the pieces as a bundle. Table~\ref{tab:priorwork} sets out that boundary: the requirement is the composition, not any single metric. We propose four changes.

\begin{table*}[t]
\centering
\small
\setlength{\tabcolsep}{6pt}
\renewcommand{\arraystretch}{1.15}
\begin{tabular}{@{}p{0.26\textwidth}p{0.32\textwidth}p{0.34\textwidth}@{}}
\toprule
\textbf{Prior work} & \textbf{Axis it covers} & \textbf{Incentive it does not create} \\
\midrule
HELM, TruthfulQA, AbstentionBench & Calibration, truthful non-commitment, direct abstention measurement & Decline is not a separately scored action on a board models are ranked on \\
TruthRL, AbstainR1, KeRLQA, Reinforced Hesitation & Ternary reward at training time & Evaluation stays binary, so the advance stays isolated \\
Kalai et al., AA-Omniscience, FACTS & Abstention-aware scoring, argued for and deployed & No capability stratification, no subgroup audit, no treatment of the pieces as a bundle \\
\midrule
\emph{This paper} & The four composed, with a $p^*$ sweep and a proper score on commitments specified & Evaluation-side only; it does not price deployment risk \\
\bottomrule
\end{tabular}
\caption{Where the proposal sits relative to prior work. Each prior row covers a real axis; none composes the four into a single requirement. The scoring rule itself is Chow's and is not claimed as new.}
\label{tab:priorwork}
\end{table*}

\begin{enumerate}
\item \textbf{Triple-scoring.} Replace the binary correct/incorrect scoring function with a three-valued outcome: correct, incorrect, or abstained, where \emph{abstain} is a distinct action the model selects rather than a missing prediction. For open-weight base models that do not emit a structured \emph{abstain} action, the action can be derived post-hoc from predictive entropy thresholding, semantic-uncertainty clustering across multiple stochastic generations \citep{kuhn2023semantic}, or conformal selection layers \citep{quach2024conformal}; the choice does not affect the design condition below. Let $p(c \mid x), p(w \mid x), p(a \mid x) \geq 0$ with $p(c \mid x) + p(w \mid x) + p(a \mid x) = 1$ denote the model's empirical probabilities of correct, incorrect, and abstain outcomes on input $x$, estimated from $k$ rollouts under the model's decoding distribution. Abstention receives positive weight on problems where the model is outside its reliable capability and zero weight elsewhere. The design condition is the reward inequality
\begin{equation}
R_{\text{abstain}} > \tilde{p}(c|x) \cdot R_{\text{correct}} + \tilde{p}(w|x) \cdot R_{\text{incorrect}}
\label{eq:reward}
\end{equation}
where $R_{\text{correct}} > 0$ is the reward for a correct attempt and the right-hand side is the expected return conditional on attempting, written with $\tilde{p}(c \mid x) = p(c \mid x)/(1 - p(a \mid x))$ and $\tilde{p}(w \mid x) = p(w \mid x)/(1 - p(a \mid x))$, which sum to one over the two attempted outcomes. The un-renormalized rates return in the scalarization below, where $p(c)$ denotes their benchmark-level mean. Equation~\ref{eq:reward} is the standard cost-sensitive abstention threshold of \citet{chow1970optimum}: it fixes the model's threshold on subjective probability at $p^* = R_{\text{abstain}} / R_{\text{correct}}$ when $R_{\text{incorrect}} = 0$. The rule is classical. What is new is applying it where the abstain action has been priced at zero. Setting $R_{\text{abstain}}/R_{\text{correct}}$ is therefore a leaderboard-design decision, not a mathematical given: it fixes $p^*$, and so the tolerance for being wrong relative to being silent. No single threshold fits every domain --- creative writing, medical QA, and legal advice price a wrong commitment very differently --- so the useful convention is to publish a sweep over $R_{\text{incorrect}}$ (equivalently, over $p^*$) rather than one point estimate, letting developers read the ranking at their own cost of failure. Because each model's behavior is fixed at evaluation time, this re-scores it rather than tracing an operating characteristic; a genuine risk-coverage curve \citep{geifman2017selective} would additionally require per-item confidence, which a reform-built benchmark should collect. Expected utility at a fixed $p^*$ is a valid total order over fixed policies for exactly that reason: it needs no matched coverage, where a coverage-matched comparison or an AURC would answer the different question of how well a model ranks its own confidence. Under current scoring ($R_{\text{abstain}} = 0$, $R_{\text{incorrect}} = 0$, $R_{\text{correct}} > 0$), the inequality strictly fails for any $\tilde{p}(c \mid x) > 0$, which is why guessing dominates over decline.

\emph{Scope.} The rule scores epistemic abstention, the model recognizing a capability gap, and not prescriptive refusal, the model executing a safety policy on a dangerous query. Folding the second into capability scoring would confound the two.

\emph{Calibration verification.} A single operating point still conflates competence with coverage. A board topped by a system that never answers is decision-theoretically right and useless as a capability signal, so two things have to sit beside it. The sweep does most of the work: a policy beats blanket abstention exactly when its accuracy on attempted items exceeds $p^*$, so a board reported across $p^*$ shows at what price of error each model stops being worth consulting, and a null policy tops one only where no model clears $p^*$. The second is a strictly proper scoring rule \citep{brier1950verification,gneiting2007strictly,savage1971elicitation} on the model's expressed confidence (Brier or log-score) over its committed answers, reported beside $U$ and the abstention rate. It cannot outrank the abstainer, since a model with no commitments has no proper score at all. Its job is to verify the calibration that the optimality of $U$ assumes. A model whose expressed confidence is not calibrated is not in fact abstaining below $p^*$, whatever its abstention rate happens to be.

\emph{Leaderboard scalarization.} Reporting three numbers does not by itself produce a ranking, and practitioners rank. The ranking Equation~\ref{eq:reward} already implies is the expected utility $U = R_{\text{correct}}\,p(c) + R_{\text{incorrect}}\,p(w) + R_{\text{abstain}}\,p(a)$, ordered at the operating point $p^* = (R_{\text{abstain}} - R_{\text{incorrect}})/(R_{\text{correct}} - R_{\text{incorrect}})$, the Chow threshold. Item~1 reads it with $R_{\text{incorrect}} = 0$, giving $p^* = R_{\text{abstain}}/R_{\text{correct}}$; Table~\ref{tab:simpleqa} reads it with $R_{\text{correct}} = 1$, $R_{\text{abstain}} = 0$ and $R_{\text{incorrect}} = -\lambda$, giving $p^* = \lambda/(\lambda+1)$. A concrete instance: $R_{\text{correct}} = +1$, $R_{\text{incorrect}} = -1$, $R_{\text{abstain}} = 0$ yields $p^* = 0.5$, or ``answer only when more likely right than wrong.'' The ranking $U$ induces is invariant under positive affine transformations of the three rewards, and $p^*$ is a complete invariant: every triple with $R_{\text{incorrect}} \le R_{\text{abstain}} < R_{\text{correct}}$ reduces to $(1, 0, p^*)$. A sweep over $p^* \in [0,1)$ is therefore exhaustive over reward triples rather than a sample of them, and $p^*{=}0$ recovers accuracy-only scoring. Calibration enters not as a third weighted term but as the verification described above, reported beside $U(p^*)$ and the abstention rate. Table~\ref{tab:simpleqa} applies the rule to published numbers.

\emph{Ground-truth requirement.} At the MMLU error rates noted earlier, a calibrated model that correctly abstains on a flawed item is penalized against the labeler's mistake under any scoring scheme. Clean labels are a prerequisite for the proposal rather than an independent concern; MMLU-Redux \citep{gema-etal-2025-mmlu} re-annotates 5,700 questions across all 57 subjects and is a drop-in replacement.

\begin{table*}[t]
\centering
\small
\setlength{\tabcolsep}{6pt}
\renewcommand{\arraystretch}{1.15}
\begin{tabular}{@{}lrrrrr@{}}
\toprule
\textbf{Model} & \textbf{Correct \%} & \textbf{Incorrect \%} & \textbf{Not-att.\ \%} & \textbf{$U$ at $p^*{=}0.5$ [rank]} & \textbf{$U$ at $p^*{\approx}0.67$ [rank]} \\
\midrule
o1-preview & 42.7 & 48.1 & 9.2 & $-5.4$ [\textbf{1}] & $-53.5$ [5] \\
GPT-4o & 38.2 & 60.8 & 1.0 & $-22.6$ [6] & $-83.4$ [6] \\
Claude-3.5-sonnet & 28.9 & 36.1 & 35.0 & $-7.2$ [2] & $-43.3$ [3] \\
Claude-3-opus & 23.5 & 36.9 & 39.6 & $-13.4$ [3] & $-50.3$ [4] \\
GPT-4o-mini & 8.6 & 90.5 & 0.9 & $-81.9$ [8] & $-172.4$ [8] \\
o1-mini & 8.1 & 63.4 & 28.5 & $-55.3$ [7] & $-118.7$ [7] \\
Claude-3-sonnet & 5.7 & 19.3 & 75.0 & $-13.6$ [4] & $-32.9$ [\textbf{1}] \\
Claude-3-haiku & 5.1 & 19.6 & 75.3 & $-14.5$ [5] & $-34.1$ [2] \\
\midrule
\emph{Always-abstain null policy} & 0.0 & 0.0 & 100.0 & $0.0$ & $0.0$ \\
\bottomrule
\end{tabular}
\caption{SimpleQA \citep{wei2024simpleqa} outcome rates (their Table~3), rows in accuracy order, re-scored as $U = \text{Correct} - \lambda \cdot \text{Incorrect}$ with $R_{\text{abstain}} = 0$ and $\lambda \in \{1, 2\}$; brackets give ranks. Accuracy order and utility order come apart: GPT-4o is second on accuracy and sixth at $p^*{=}0.5$, below Claude-3-haiku, which attempts a quarter of the items. At $p^*{\approx}0.67$ the two most abstention-heavy models take the top two places and o1-preview falls to fifth. Ranks 3--5 at $p^*{=}0.5$ span 1.1 points and should be read as a tie. The null row sits above every model at both points because the best attempted-accuracy here is o1-preview's 47.0\%; four pass below $p^*{=}0.38$. SimpleQA cannot test Item~1's calibration verification, reporting no per-model proper score, and its ``not-attempted'' is a grader label on free text rather than a structured abstain action.}
\label{tab:simpleqa}
\end{table*}

\item \textbf{Abstention rate as a primary metric.} Papers should report abstention rates alongside accuracy, broken out by problem stratum. A model at 0.85 accuracy with 5\% abstention is doing something different from one at 0.85 accuracy with 40\% abstention, and leaderboards currently cannot tell us which we have.

\item \textbf{Capability-stratified evaluation.} Benchmark construction should identify problem classes where abstention is the correct behavior. Two kinds of stratification are possible. Environment-level unanswerability, as in \citet{yin2025reasoning}, is a ground-truth fact about the task: ``what is the current time in Park Forest Village'' given no tool access has no correct completion, and the benchmark can score refusal accordingly. Model-level unanswerability, as in factual QA where the answer depends on whether the model has the fact, is harder, because what counts as ``outside capability'' is model-dependent. The environment-level case is tractable now and should be adopted. The model-level case is an open problem for benchmark design. Acknowledging the distinction, rather than collapsing it, is what separates a usable proposal from a wishlist. Four reporting axes follow. Environment-level unanswerability is scorable today. A capability boundary defined by confidence crossing $p^*$ inherits the model-level problem above, so the near-term substitute is observable difficulty or subject bins, with $k$-rollout estimates reported with intervals wherever bins are close enough to change a rank. Domain and risk class matter because a different actual-use $R_{\text{incorrect}}$ implies a different $p^*$, and so a different appropriate abstention rate. Subgroup is the fourth, for the fairness audit below.

\item \textbf{Calibration metrics as mandatory companions.} Expected calibration error (ECE) and Brier score should be reported alongside accuracy for any benchmark with a probabilistic interpretation. ECE has known pathologies \citep{kumar2019verified,roelofs2022mitigating} --- bin count and binning scheme materially change the value, and the standard top-1 estimator misses class-conditional miscalibration --- so leaderboards should specify the estimator (equal-mass, equal-width, or kernel-based) or report robustness across binning schemes; Brier and log-score are strictly proper alternatives that avoid the binning question entirely, at the cost of unboundedness on zero-probability predictions for log-score. The structural need is metric-independent: a model correct 70\% of the time at uniform 0.95 confidence is over-confident, while a model correct 50\% of the time at uniform 0.5 confidence is well-calibrated despite lower accuracy, and only the second is safe to deploy where the actual-use $R_{\text{incorrect}}$ is sharply negative. Accuracy alone cannot distinguish the two. On open-ended generation tasks (HumanEval, GSM8K), the natural probability over outputs is not directly available, and calibration must be measured on derived quantities --- semantic-equivalence-cluster probabilities \citep{kuhn2023semantic}, sampling-based confidence estimates, or LLM-as-judge graded confidence --- whose choice is consequential and must be specified per benchmark. A proper score over committed answers is computed on a self-selected subsample, and coverage varies widely across models: in Table~\ref{tab:simpleqa} the not-attempted rate ranges from 0.9\% to 75.3\%. Calibration figures are therefore not comparable across models without coverage reported beside them, and are strictly comparable only at matched coverage, since a model that commits only where it is confident will look better calibrated for that reason alone.
\end{enumerate}

The four changes are complementary. Triple-scoring without abstention-rate reporting cannot be interpreted across models; abstention-rate reporting without triple-scoring is description without incentive; capability-stratified evaluation without triple-scoring has no slice on which to differentially reward refusal; and calibration metrics without abstention reporting cannot distinguish over-confidence from miscalibrated coverage. We neither claim the bundle is minimal nor prove it sufficient; both stay open. What we do observe is that weaker bundles already adopted (HELM, AbstentionBench) have not produced the leaderboard-level gradient the proposal requires.

None of these changes requires new mathematical infrastructure. They require that the scoring function stop rewarding the failure mode all three findings describe.

\paragraph{Fairness audit.} The proposal requires a fairness audit. Triple-scoring incentivizes abstention, and abstention is not symmetric across populations: a model that abstains on Swahili medical QA because it lacks competence is correctly rewarded, but the resulting abstention-rate disparity --- between high-resource and low-resource languages, where existing multilingual leaderboards already supply the axes, between training-data-represented and underrepresented groups --- needs to be reported alongside aggregate abstention rates. Without a stratified abstention-rate audit, triple-scoring converts a population-level information asymmetry into a hidden one. A higher stratified abstention rate is appropriate when it tracks a capability limit on that stratum, and undesirable when items of equal difficulty draw systematically more abstention for one group. Pooled, the two are indistinguishable. We therefore recommend reporting accuracy-given-attempted on the same strata as the abstention rate, matched for item difficulty. Accuracy-given-attempted is what separates the benign case from the harmful one.

\paragraph{Transition path.} The lowest-cost path to adoption mirrors HELM's playbook: add triple-scoring and stratified abstention rate as side metrics on existing benchmarks, demonstrate that the resulting rankings differ from accuracy-only rankings in load-bearing ways, and only then promote them to primary. Three costs are real: annotating environment-level unanswerability, fixing the operating point, and losing comparability with historical accuracy-only scores, which a parallel-reporting transition cycle absorbs. One asymmetry deserves naming: triple-scoring requires that models emit a structured \emph{abstain} action and that benchmarks recognize it. Proprietary models can be retrained with refusal heads and calibration targets, while open-source releases are typically evaluated post-hoc against a fixed prompt template --- so a leaderboard that requires structured abstention will systematically advantage proprietary releases unless the benchmark itself supplies the prompt-template policy that elicits the abstain action.

\section{Counterexamples and Boundary Conditions}
\label{sec:counter}

The strongest objection to the position taken here is that benchmark reform is plainly not a prerequisite for reliability progress, since several deployed model families --- including those produced by Anthropic, OpenAI, and Google --- have measurably improved abstention behavior through honesty-oriented post-training \citep{yang2024alignment} (refusal training, constitutional methods, calibration heads) without any of the benchmark changes proposed in Section~\ref{sec:proposal}. We take the objection seriously. If post-training closes the gap the theorem identifies, the four reforms are optional optimizations rather than structural requirements.

The empirical evidence does not support that reading, but it does support a weaker version of it. \citet{kirichenko2025abstentionbench} measure a 24\% mean degradation in abstention across their 20 datasets, over the two reasoning and non-reasoning model pairs they compare under reasoning fine-tuning, including in domains the reasoning models were explicitly trained on. \citet{yin2025reasoning}'s reliability-capability trade-off persists across the mitigation strategies they test, with DPO reducing hallucination only by sacrificing utility. Honesty post-training narrows the gap on the slice of inputs where the model itself recognizes uncertainty. We conjecture --- and treat this as the testable empirical claim of the position rather than as an established fact --- that it does not narrow the gap on the complementary slice the theorem singles out: inputs where consistent reasoning would force confabulation in a region the model believes it knows. The gradients addressing them differ, and the second is the one current benchmarks suppress.

To anchor this empirically, we name a falsification condition: if a post-training intervention without scoring reform achieved sub-10\% residual error on a calibrated abstention benchmark stratified by environment-level unanswerability --- at the level AbstentionBench currently measures --- we would withdraw the structural-necessity reading and treat benchmark reform as a complement rather than a prerequisite for principled closure of the theorem-identified class. Environment-level unanswerability is a proxy here: the slice the conjecture names is model-level, which Item~3 marks as open, so this is the strongest test currently runnable rather than the exact one.

We therefore do not claim that post-training is mistaken or that the field should abandon it. The narrower position is that benchmark reform is necessary, not sufficient, for closing the class Bastounis identifies, and post-training is necessary, not sufficient, for closing the class users most often encounter. The two interventions address different gradients, and a reliability program that pursues only one will plateau at the boundary of the other.

\section*{Limitations}

We do not claim the convergence implies that the three papers' specific interventions or proposed mitigations are equivalent. Yin's work points at activation-level diagnostics. Suleymanov's points at agentic scaffolding for refinement. Bastounis's points at the ``I don't know'' function as a recursion-theoretic object. These remain distinct research programs even if the warranted intervention is the same.

We do not claim our four-part evaluation proposal is the only possible response to the diagnosis. A reasonable alternative would propose mechanism-level interventions (calibration heads, refusal tokens, conformal layers) that retrofit existing benchmarks rather than reforming them. We argue that without scoring reform such interventions are selected against rather than for.

We do not claim our convergence reading is the only possible reading of Yin, Suleymanov, and Bastounis. Other accounts may fit the three papers equally well. We claim that treating them as three separate reliability problems misses what they share, and that what they share has policy consequences the field has not yet acted on.

We do not claim the three findings represent the same kind of unavailability. Yin's is a capability gap; Suleymanov's a policy gap; Bastounis's theorem is agnostic to source --- which is what lets the three findings converge on the same remedy despite having different origins. The distinction matters for evaluation design: capability gaps can be probed with tasks that are ground-truth unanswerable, policy gaps require benchmarks that specify the ruling-out rule. Our Section~\ref{sec:proposal} proposal scores the first and leaves the second to the safety benchmarks that already score it, which is the separation Table~\ref{tab:taxonomy} draws.

We do not claim the proposal generalizes beyond discrete-answer settings. For long-form generation, and equally for multi-step and interactive settings where a commitment is distributed across turns rather than made once, calibrated abstention has no clean discrete analogue at the document or dialogue level, and operationalizing triple-scoring there is an open problem outside this proposal's scope. Much deployed unreliability lives in exactly those settings, so we mark them as the reform's principal open extension rather than as ground it already covers.

\section*{Use of AI Assistants}

AI assistants were used to support drafting, editing, and structural revision of the
manuscript text, and to assist with \LaTeX{} formatting and bibliography verification.
The author conceived the central argument (the convergence claim and the four-part
benchmark reform), selected and read all cited works, made all substantive claims, and
verified all citations against primary sources. All factual and theoretical assertions in
the paper are the author's responsibility. AI assistants were not used to generate
citations or fabricate evidence.

\bibliography{references}

\begin{thebibliography}{35}
\providecommand{\natexlab}[1]{#1}

\bibitem[{Bastounis et~al.(2024)Bastounis, Campodonico, van~der Schaar, Adcock,
  and Hansen}]{bastounis2024consistent}
Alexander Bastounis, Paolo Campodonico, Mihaela van~der Schaar, Ben Adcock, and
  Anders~C. Hansen. 2024.
\newblock \href {https://arxiv.org/abs/2408.02357} {On the consistent reasoning
  paradox of intelligence and optimal trust in ai: The power of 'i don't
  know'}.
\newblock \emph{Preprint}, arXiv:2408.02357.

\bibitem[{Brier(1950)}]{brier1950verification}
Glenn~W. Brier. 1950.
\newblock Verification of forecasts expressed in terms of probability.
\newblock \emph{Monthly Weather Review}, 78(1):1--3.

\bibitem[{Chen et~al.(2021)Chen, Tworek, Jun et~al.}]{chen2021humaneval}
Mark Chen, Jerry Tworek, Heewoo Jun, et~al. 2021.
\newblock \href {https://arxiv.org/abs/2107.03374} {Evaluating large language
  models trained on code}.
\newblock \emph{Preprint}, arXiv:2107.03374.

\bibitem[{Cheng et~al.(2025)Cheng, Jacovi, Globerson et~al.}]{cheng2025facts}
Aileen Cheng, Alon Jacovi, Amir Globerson, et~al. 2025.
\newblock \href {https://arxiv.org/abs/2512.10791} {The {FACTS} leaderboard: A
  comprehensive benchmark for large language model factuality}.
\newblock \emph{Preprint}, arXiv:2512.10791.

\bibitem[{Chow(1970)}]{chow1970optimum}
C.~K. Chow. 1970.
\newblock \href {https://doi.org/10.1109/TIT.1970.1054406} {On optimum
  recognition error and reject tradeoff}.
\newblock \emph{{IEEE} Transactions on Information Theory}, 16(1):41--46.

\bibitem[{Cobbe et~al.(2021)Cobbe, Kosaraju, Bavarian, Chen, Jun, Kaiser,
  Plappert, Tworek, Hilton, Nakano, Hesse, and Schulman}]{cobbe2021gsm8k}
Karl Cobbe, Vineet Kosaraju, Mohammad Bavarian, Mark Chen, Heewoo Jun, Lukasz
  Kaiser, Matthias Plappert, Jerry Tworek, Jacob Hilton, Reiichiro Nakano,
  Christopher Hesse, and John Schulman. 2021.
\newblock \href {https://arxiv.org/abs/2110.14168} {Training verifiers to solve
  math word problems}.
\newblock \emph{Preprint}, arXiv:2110.14168.

\bibitem[{El-Yaniv and Wiener(2010)}]{el2010foundations}
Ran El-Yaniv and Yair Wiener. 2010.
\newblock On the foundations of noise-free selective classification.
\newblock \emph{Journal of Machine Learning Research}, 11:1605--1641.

\bibitem[{Geifman and El-Yaniv(2017)}]{geifman2017selective}
Yonatan Geifman and Ran El-Yaniv. 2017.
\newblock Selective classification for deep neural networks.
\newblock In \emph{Advances in Neural Information Processing Systems 30
  (NeurIPS)}.

\bibitem[{Gema et~al.(2025)Gema, Leang, Hong, Devoto, Mancino, Saxena, He,
  Zhao, Du, Ghasemi~Madani, Barale, McHardy, Harris, Kaddour, Van~Krieken, and
  Minervini}]{gema-etal-2025-mmlu}
Aryo~Pradipta Gema, Joshua Ong~Jun Leang, Giwon Hong, Alessio Devoto, Alberto
  Carlo~Maria Mancino, Rohit Saxena, Xuanli He, Yu~Zhao, Xiaotang Du,
  Mohammad~Reza Ghasemi~Madani, Claire Barale, Robert McHardy, Joshua Harris,
  Jean Kaddour, Emile Van~Krieken, and Pasquale Minervini. 2025.
\newblock \href {https://doi.org/10.18653/v1/2025.naacl-long.262} {Are we done
  with {MMLU}?}
\newblock In \emph{Proceedings of the 2025 Conference of the Nations of the
  Americas Chapter of the Association for Computational Linguistics: Human
  Language Technologies (Volume 1: Long Papers)}, pages 5069--5096,
  Albuquerque, New Mexico. Association for Computational Linguistics.

\bibitem[{Gneiting and Raftery(2007)}]{gneiting2007strictly}
Tilmann Gneiting and Adrian~E. Raftery. 2007.
\newblock \href {https://doi.org/10.1198/016214506000001437} {Strictly proper
  scoring rules, prediction, and estimation}.
\newblock \emph{Journal of the American Statistical Association},
  102(477):359--378.

\bibitem[{Hendrycks et~al.(2021)Hendrycks, Burns, Basart, Zou, Mazeika, Song,
  and Steinhardt}]{hendrycks2021measuring}
Dan Hendrycks, Collin Burns, Steven Basart, Andy Zou, Mantas Mazeika, Dawn
  Song, and Jacob Steinhardt. 2021.
\newblock Measuring massive multitask language understanding.
\newblock In \emph{International Conference on Learning Representations
  (ICLR)}.
\newblock ArXiv:2009.03300.

\bibitem[{Jackson et~al.(2025)Jackson, Keating, Cameron, and
  Hill-Smith}]{jackson2025omniscience}
Declan Jackson, William Keating, George Cameron, and Micah Hill-Smith. 2025.
\newblock \href {https://arxiv.org/abs/2511.13029} {{AA-Omniscience}:
  Evaluating cross-domain knowledge reliability in large language models}.
\newblock \emph{Preprint}, arXiv:2511.13029.

\bibitem[{Kadavath et~al.(2022)Kadavath, Conerly, Askell, Henighan, Drain,
  Perez, Schiefer, Hatfield-Dodds, DasSarma, Tran-Johnson, Johnston, El-Showk,
  Jones, Elhage, Hume, Chen, Bai, Bowman, Fort, Ganguli, Hernandez, Jacobson,
  Kernion, Kravec, Lovitt, Ndousse, Olsson, Ringer, Amodei, Brown, Clark,
  Joseph, Mann, McCandlish, Olah, and Kaplan}]{kadavath2022language}
Saurav Kadavath, Tom Conerly, Amanda Askell, Tom Henighan, Dawn Drain, Ethan
  Perez, Nicholas Schiefer, Zac Hatfield-Dodds, Nova DasSarma, Eli
  Tran-Johnson, Scott Johnston, Sheer El-Showk, Andy Jones, Nelson Elhage,
  Tristan Hume, Anna Chen, Yuntao Bai, Sam Bowman, Stanislav Fort, and 17
  others. 2022.
\newblock Language models (mostly) know what they know.
\newblock \emph{arXiv preprint arXiv:2207.05221}.

\bibitem[{Kalai et~al.(2026)Kalai, Nachum, Vempala, and
  Zhang}]{kalai2025hallucinate}
Adam~Tauman Kalai, Ofir Nachum, Santosh~S. Vempala, and Edwin Zhang. 2026.
\newblock \href {https://doi.org/10.1038/s41586-026-10549-w} {Evaluating large
  language models for accuracy incentivizes hallucinations}.
\newblock \emph{Nature}.
\newblock Published 22 April 2026; preprint arXiv:2509.04664, ``Why Language
  Models Hallucinate''.

\bibitem[{Kamath et~al.(2020)Kamath, Jia, and
  Liang}]{kamath-etal-2020-selective}
Amita Kamath, Robin Jia, and Percy Liang. 2020.
\newblock \href {https://doi.org/10.18653/v1/2020.acl-main.503} {Selective
  question answering under domain shift}.
\newblock In \emph{Proceedings of the 58th Annual Meeting of the Association
  for Computational Linguistics}, pages 5684--5696. Association for
  Computational Linguistics.

\bibitem[{Kirichenko et~al.(2025)Kirichenko, Ibrahim, Chaudhuri, and
  Bell}]{kirichenko2025abstentionbench}
Polina Kirichenko, Mark Ibrahim, Kamalika Chaudhuri, and Samuel~J. Bell. 2025.
\newblock \href {https://openreview.net/forum?id=OkHC30LLpO}
  {{AbstentionBench}: Reasoning {LLMs} fail on unanswerable questions}.
\newblock In \emph{Advances in Neural Information Processing Systems 38:
  Datasets and Benchmarks Track}.
\newblock ArXiv:2506.09038.

\bibitem[{Kuhn et~al.(2023)Kuhn, Gal, and Farquhar}]{kuhn2023semantic}
Lorenz Kuhn, Yarin Gal, and Sebastian Farquhar. 2023.
\newblock Semantic uncertainty: Linguistic invariances for uncertainty
  estimation in natural language generation.
\newblock In \emph{The Eleventh International Conference on Learning
  Representations (ICLR)}.

\bibitem[{Kumar et~al.(2019)Kumar, Liang, and Ma}]{kumar2019verified}
Ananya Kumar, Percy~S. Liang, and Tengyu Ma. 2019.
\newblock Verified uncertainty calibration.
\newblock In \emph{Advances in Neural Information Processing Systems
  (NeurIPS)}.

\bibitem[{Liang et~al.(2023)Liang, Bommasani, Lee, Tsipras, Soylu, Yasunaga,
  Zhang, Narayanan, Wu, Kumar, Newman, Yuan, Yan, Zhang, Cosgrove, Manning,
  R{\'e}, Acosta-Navas, Hudson, Zelikman, Durmus, Ladhak, Rong, Ren, Yao, Wang,
  Santhanam, Orr, Zheng, Yuksekgonul, Suzgun, Kim, Guha, Chatterji, Khattab,
  Henderson, Huang, Chi, Xie, Santurkar, Ganguli, Hashimoto, Icard, Zhang,
  Chaudhary, Wang, Li, Mai, Zhang, and Koreeda}]{liang2022holistic}
Percy Liang, Rishi Bommasani, Tony Lee, Dimitris Tsipras, Dilara Soylu,
  Michihiro Yasunaga, Yian Zhang, Deepak Narayanan, Yuhuai Wu, Ananya Kumar,
  Benjamin Newman, Binhang Yuan, Bobby Yan, Ce~Zhang, Christian Cosgrove,
  Christopher~D. Manning, Christopher R{\'e}, Diana Acosta-Navas, Drew~A.
  Hudson, and 31 others. 2023.
\newblock Holistic evaluation of language models.
\newblock \emph{Transactions on Machine Learning Research (TMLR)}.
\newblock ArXiv:2211.09110, originally posted Nov 2022.

\bibitem[{Lin et~al.(2022)Lin, Hilton, and Evans}]{lin2022truthfulqa}
Stephanie Lin, Jacob Hilton, and Owain Evans. 2022.
\newblock \href {https://doi.org/10.18653/v1/2022.acl-long.229} {{TruthfulQA}:
  Measuring how models mimic human falsehoods}.
\newblock In \emph{Proceedings of the 60th Annual Meeting of the Association
  for Computational Linguistics (ACL)}, pages 3214--3252. Association for
  Computational Linguistics.

\bibitem[{Min et~al.(2023)Min, Krishna, Lyu, Lewis, Yih, Koh, Iyyer,
  Zettlemoyer, and Hajishirzi}]{min2023factscore}
Sewon Min, Kalpesh Krishna, Xinxi Lyu, Mike Lewis, Wen-tau Yih, Pang~Wei Koh,
  Mohit Iyyer, Luke Zettlemoyer, and Hannaneh Hajishirzi. 2023.
\newblock \href {https://doi.org/10.18653/v1/2023.emnlp-main.741} {{FActScore}:
  Fine-grained atomic evaluation of factual precision in long form text
  generation}.
\newblock In \emph{Proceedings of the 2023 Conference on Empirical Methods in
  Natural Language Processing (EMNLP)}.
\newblock ArXiv:2305.14251.

\bibitem[{Mohamadi et~al.(2025)Mohamadi, Wang, and Li}]{mohamadi2025hesitation}
Mohamad~Amin Mohamadi, Tianhao Wang, and Zhiyuan Li. 2025.
\newblock \href {https://arxiv.org/abs/2511.11500} {Honesty over accuracy:
  Trustworthy language models through reinforced hesitation}.
\newblock \emph{Preprint}, arXiv:2511.11500.

\bibitem[{Quach et~al.(2024)Quach, Fisch, Schuster, Yala, Sohn, Jaakkola, and
  Barzilay}]{quach2024conformal}
Victor Quach, Adam Fisch, Tal Schuster, Adam Yala, Jae~Ho Sohn, Tommi~S.
  Jaakkola, and Regina Barzilay. 2024.
\newblock \href {https://openreview.net/forum?id=pzUhfQ74c5} {Conformal
  language modeling}.
\newblock In \emph{The Twelfth International Conference on Learning
  Representations (ICLR)}.

\bibitem[{Ralethe and Buys(2025)}]{ralethe2025kerlqa}
Sello Ralethe and Jan Buys. 2025.
\newblock \href {https://doi.org/10.18653/v1/2025.ijcnlp-long.99} {{KERLQA}:
  Knowledge-enhanced reinforcement learning for question answering in
  low-resource languages}.
\newblock In \emph{Proceedings of the 14th International Joint Conference on
  Natural Language Processing and the 4th Conference of the Asia-Pacific
  Chapter of the Association for Computational Linguistics}, pages 1834--1846,
  Mumbai, India. The Asian Federation of Natural Language Processing and The
  Association for Computational Linguistics.

\bibitem[{Roelofs et~al.(2022)Roelofs, Cain, Shlens, and
  Mozer}]{roelofs2022mitigating}
Rebecca Roelofs, Nicholas Cain, Jonathon Shlens, and Michael~C. Mozer. 2022.
\newblock Mitigating bias in calibration error estimation.
\newblock In \emph{Proceedings of the International Conference on Artificial
  Intelligence and Statistics (AISTATS)}.

\bibitem[{Savage(1971)}]{savage1971elicitation}
Leonard~J. Savage. 1971.
\newblock \href {https://doi.org/10.1080/01621459.1971.10482346} {Elicitation
  of personal probabilities and expectations}.
\newblock \emph{Journal of the American Statistical Association},
  66(336):783--801.

\bibitem[{Suleymanov et~al.(2026)Suleymanov, Rajabov, Mirzazada, and
  Kantarcioglu}]{suleymanov2026beyond}
Umid Suleymanov, Zaur Rajabov, Emil Mirzazada, and Murat Kantarcioglu. 2026.
\newblock \href {https://arxiv.org/abs/2602.21496} {Beyond refusal: Probing the
  limits of agentic self-correction for semantic sensitive information}.
\newblock \emph{Preprint}, arXiv:2602.21496.

\bibitem[{Wei et~al.(2024)Wei, Karina, Chung, Jiao, Papay, Glaese, Schulman,
  and Fedus}]{wei2024simpleqa}
Jason Wei, Nguyen Karina, Hyung~Won Chung, Yunxin~Joy Jiao, Spencer Papay,
  Amelia Glaese, John Schulman, and William Fedus. 2024.
\newblock \href {https://arxiv.org/abs/2411.04368} {Measuring short-form
  factuality in large language models}.
\newblock \emph{Preprint}, arXiv:2411.04368.

\bibitem[{Wei et~al.(2025)Wei, Yang, Sun, Wang, Shao, Chen, Kachuee, Gollapudi,
  Liao, Scheffer, Wanga, Kumar, Meng, tau Yih, and Dong}]{wei2025truthrl}
Zhepei Wei, Xiao Yang, Kai Sun, Jiaqi Wang, Rulin Shao, Jingxiang Chen,
  Mohammad Kachuee, Teja Gollapudi, Yiwei Liao, Nicolas Scheffer, Rakesh Wanga,
  Anuj Kumar, Yu~Meng, Wen tau Yih, and Xin~Luna Dong. 2025.
\newblock \href {https://arxiv.org/abs/2509.25760} {{TruthRL}: Incentivizing
  truthful {LLMs} via reinforcement learning}.
\newblock \emph{Preprint}, arXiv:2509.25760.
\newblock ICML 2026.

\bibitem[{Wen et~al.(2025)Wen, Yao, Feng, Xu, Tsvetkov, Howe, and
  Wang}]{wen-etal-2025-know}
Bingbing Wen, Jihan Yao, Shangbin Feng, Chenjun Xu, Yulia Tsvetkov, Bill Howe,
  and Lucy~Lu Wang. 2025.
\newblock \href {https://doi.org/10.1162/tacl_a_00754} {Know your limits: A
  survey of abstention in large language models}.
\newblock \emph{Transactions of the Association for Computational Linguistics},
  13:529--556.
\newblock ArXiv:2407.18418.

\bibitem[{Yang et~al.(2024)Yang, Chern, Qiu, Neubig, and
  Liu}]{yang2024alignment}
Yuqing Yang, Ethan Chern, Xipeng Qiu, Graham Neubig, and Pengfei Liu. 2024.
\newblock Alignment for honesty.
\newblock In \emph{Proceedings of the 38th International Conference on Neural
  Information Processing Systems (NeurIPS '24)}, Red Hook, NY, USA. Curran
  Associates Inc.

\bibitem[{Yin et~al.(2026)Yin, Sha, Cui, Meng, and Li}]{yin2025reasoning}
Chenlong Yin, Zeyang Sha, Shiwen Cui, Changhua Meng, and Zechao Li. 2026.
\newblock \href {https://doi.org/10.18653/v1/2026.acl-long.376} {The reasoning
  trap: How enhancing {LLM} reasoning amplifies tool hallucination}.
\newblock In \emph{Annual Meeting of the Association for Computational
  Linguistics (ACL)}.
\newblock ArXiv:2510.22977; the arXiv record states ``Accepted to ACL 2026
  Main''.

\bibitem[{Zellers et~al.(2019)Zellers, Holtzman, Bisk, Farhadi, and
  Choi}]{zellers2019hellaswag}
Rowan Zellers, Ari Holtzman, Yonatan Bisk, Ali Farhadi, and Yejin Choi. 2019.
\newblock \href {https://doi.org/10.18653/v1/P19-1472} {{HellaSwag}: Can a
  machine really finish your sentence?}
\newblock In \emph{Proceedings of the 57th Annual Meeting of the Association
  for Computational Linguistics}.
\newblock ArXiv:1905.07830.

\bibitem[{Zhai et~al.(2026)Zhai, Liang, and Kang}]{zhai2026abstainr1}
Skylar Zhai, Jingcheng Liang, and Dongyeop Kang. 2026.
\newblock \href {https://arxiv.org/abs/2604.17073} {{Abstain-R1}: Calibrated
  abstention and post-refusal clarification via verifiable {RL}}.
\newblock \emph{Preprint}, arXiv:2604.17073.

\bibitem[{Zhang et~al.(2024)Zhang, Diao, Lin, Fung, Lian, Wang, Chen, Ji, and
  Zhang}]{zhang2024rtuning}
Hanning Zhang, Shizhe Diao, Yong Lin, Yi~Fung, Qing Lian, Xingyao Wang, Yangyi
  Chen, Heng Ji, and Tong Zhang. 2024.
\newblock \href {https://doi.org/10.18653/v1/2024.naacl-long.394} {{R}-tuning:
  Instructing large language models to say `{I} don{'}t know'}.
\newblock In \emph{Proceedings of the 2024 Conference of the North American
  Chapter of the Association for Computational Linguistics: Human Language
  Technologies (Volume 1: Long Papers)}, pages 7113--7139, Mexico City, Mexico.
  Association for Computational Linguistics.

\end{thebibliography}

\end{document}